\documentclass[letterpaper]{article} % DO NOT CHANGE THIS
\usepackage[]{aaai23}  % DO NOT CHANGE THIS
\usepackage{times}  % DO NOT CHANGE THIS
\usepackage{helvet}  % DO NOT CHANGE THIS
\usepackage{courier}  % DO NOT CHANGE THIS
\usepackage[hyphens]{url}  % DO NOT CHANGE THIS
\usepackage{graphicx} % DO NOT CHANGE THIS
\usepackage{natbib}  % DO NOT CHANGE THIS AND DO NOT ADD ANY OPTIONS TO IT
\usepackage{caption,subcaption} % DO NOT CHANGE THIS AND DO NOT ADD ANY OPTIONS TO IT
\DeclareCaptionStyle{ruled}{labelfont=normalfont,labelsep=colon,strut=off} % DO NOT CHANGE THIS
\usepackage{algorithm}
\usepackage{algorithmic,booktabs,subcaption}

\usepackage{amsfonts,amsmath,bm}
\usepackage{color,colortbl}

\definecolor{Gray}{gray}{0.9}

\newcommand{\expected}{\mathbb{E}}
\newcommand{\bmx}{\bm{x}}
\newcommand{\bmz}{\bm{z}}

\newcommand{\yfact}{y_{\texttt{fact}}}
\newcommand{\yfoil}{y_{\texttt{contrast}}}

\newcommand{\method}{CEnt}

\usepackage{inconsolata}
\usepackage{amssymb}% http://ctan.org/pkg/amssymb
\usepackage{pifont}% http://ctan.org/pkg/pifont
\newcommand{\cmark}{\ding{51}}%
\newcommand{\xmark}{\ding{55}}%

\usepackage{newfloat}
\usepackage{listings}
\floatstyle{ruled}
\newfloat{listing}{tb}{lst}{}
\floatname{listing}{Listing}
\nocopyright

\title{CoTE: A Novel Explainability Method to Contrast the Decision of \textit{any} Classifier}
\title{CEnt: A Novel Explainability Method to \textit{C}ontrast the Decision of any Classifier Based on \textit{Ent}ropy Information}
\title{A Novel Explainability Method to Contrast the Decision of \textit{any} Classifier Based on Entropy Information}
\title{CEnt: An Entropy-based Model-agnostic Explainability Framework to Contrast Classifiers’ Decisions}
\author{
    Julia El Zini\equalcontrib, Mohammad Mansour\equalcontrib and Mariette Awad
}
\affiliations{Electrical and Computer Engineering Department\\ American University of Beirut\\
    Beirut, Lebanon\\
    \{jwe04,mgm35\}@mail.aub.edu and mariette.awad@aub.edu.lb
}

\begin{document}

\maketitle

\begin{abstract}
 Current interpretability methods focus on explaining a particular model's decision through present input features. Such methods do not inform the user of the sufficient conditions that alter these decisions when they are not desirable. Contrastive explanations circumvent this problem by providing explanations of the form ``If the feature $X>x$, the output $Y$ would be different''. While different approaches are developed to find proximate counterfactual examples that would alter the model's prediction; these methods do not all deal with mutability and attainability constraints.

\textit{Rather than explaining the reason behind an unfavorable outcome, teach me how to achieve a favorable one.} In this work, we
present a novel approach to locally contrast the prediction of \textit{any} classifier. Our \textbf{C}ontrastive \textbf{Ent}ropy-based explanation method, \textbf{\method{}}, approximates a model locally by a decision tree to compute entropy information of different feature splits. A graph $\mathcal{G}$ is then built where contrast nodes are found through a \textit{one-to-many} shortest path search. Contrastive examples are generated from the shortest path to reflect feature splits that alter model decision while maintaining lower entropy. We perform local sampling on manifold-like distances computed by variational auto-encoders to reflect data density. \method{} is the first non-gradient based contrastive method generating diverse counterfactuals that do not necessarily exist in the training data while satisfying immutability (ex. race) and semi-immutability (ex. age can only change in an increasing direction). Empirical evaluation on four real-world numerical datasets demonstrates the ability of \method{} in generating counterfactuals that achieve better proximity rates than existing methods without compromising latency, feasibility and attainability\footnote{Code available on https://github.com/mohamadmansourX/CEnt}. We further extend \method{} to imagery data to derive visually appealing and useful contrasts between class labels on MNIST and Fashion MNIST datasets. Finally, we show how \method{} can serve as a tool to detect vulnerabilities of textual classifiers. 

\end{abstract}

\section{Introduction}\label{sec:intro}
Ethical considerations in AI have driven rapid progress in Explainable AI (ExAI) methods especially in high stake areas such as healthcare and criminology. Existing ExAI methods 
% rely on input perturbation \cite{zeiler2014visualizing,datta2015influence}, saliency and visualization techniques \cite{simonyan2013deep} \cite{dosovitskiy2015inverting,zhou2016learning} and ideas borrowed from game theory \cite{lundberg2017unified} to explain the decision of machine learning (ML) models. These methods 
look for input features, or set of features, that influence the model's decision \cite{zeiler2014visualizing,zhou2016learning,lundberg2017unified}. However, practitioners and data subjects pose strong requirements on the \textit{usefulness} aspect of explanations which implies a selective, contrastive and social process \cite{are_explanations_useful,ribera2019can,mittelstadt2019explaining}. This aspect entails an actionable plan which serves as a \textit{constructive feedback} when the prediction is not favorable. As an example, a loan applicant is more interested in how to get their applications accepted rather than the \textit{reason} behind the rejection \cite{wachter2017counterfactual}. 

To this end, ExAI is witnessing a new vein of methods explaining decisions through contrastive learning motivated by the seminal work of \cite{wachter2017counterfactual}. These recourse (a.k.a counterfactual) methods search for a proximal input that can alter the prediction. They often employ causal graphs, gradient-descent, discriminative and evolutionary algorithms to generate contrastive examples (CEs) while satisfying feasibility constraints \cite{ustun2019actionable,o2020generative,goyal2019explaining}. Apart from being able to contrast the outputs of various models, the most wanted desiderata of counterfactuals are plausibility, attainability and diversity. While contrasting the output is successfully achieved by all methods, other requirements partake a trade-off and are rarely simultaneously satisfied. For instance, some methods violate constraints \cite{wachter2017counterfactual}, others do not always yield attainable counterfactuals \cite{mothilal2020dice} or output a unique CE based on a proximity measure \cite{dhurandhar2018cem}. 

However, the daunting acquisition of the causal graphs and the unavailability of user-defined similarity measures limit the adoption of such techniques in practice \cite{joshi2019towards}. Furthermore, gradient-based methods are sensitive to the classification boundary and the geometry of the underlying data distribution \cite{downs2020cruds}. Operating on features that are \textit{present} in the input even if their perturbation might yield explanations that are negatively contributing to the decision-making process \cite{dhurandhar2018explanations}. More importantly, existing methods prevent downstream users from exploring alternatives and specifying cost for alterations in an ad-hoc manner.

In this work, we address the shortcomings of existing methods and design a \textbf{C}ontrastive \textbf{Ent}ropy-based Explainability method, \method{}, under feasibility, immutability and semi-immutability constraints while satisfying proximity and user-defined costs. Given an observation $x$, \method{} samples $k$ local neighbors of $x$ based on manifold-like distance approximated by Variational Auto-Encoders (VAEs). Then, \method{} approximates a black-box machine learning model by a decision tree in the local neighborhood. A graph is built on top of the trained tree via a carefully-designed edge weighting scheme that compactly integrates the constraints. A one-to-many graph search technique then serves as a diverse counterfactual generation scheme in low-entropy decision sub-spaces. \method{} is the \textit{first} \textit{model-agnostic} recourse method that does not pose differentiable requirements on the black-box model and satisfies immutability and semi-immutability constraints. It can also deal with categorical data and generates diverse counterfactuals that are attainable according to the underlying data distribution while allowing for user-defined feature costs. Our validation demonstrates the effectiveness of \method{} as it yields proximate counterfactuals while achieving low latency and constraint violation rates and high attainability. Our extension to imagery data and convolutional neural networks (CNNs) shows that \method{} can derive visual contrasts that are minimal and more useful than traditional explainability methods such as LIME \cite{ribeiro2016should}. Lastly, \method{} can highlight weaknesses of textual classifiers by deriving adversarial attacks.

Next, we survey existing recourse methods and motivate \method{} before describing our methodology and studying its complexity. We then validate \method{} on different data types and model architectures before highlighting the limitations when concluding our work.

\section{Related Work}\label{sec:lit}
\begin{table*}[t]
\footnotesize
\begin{tabular}{lcccccc}
  & Model- & Immutability & Categorical & Semi- & Diversity         & Generated \\
    & agnostic &  &  & immutability &          &  \\\hline\hline
CEM  \cite{dhurandhar2018cem}   & \xmark             & \xmark           & \xmark          & \xmark                & \xmark                & \cmark          \\
AR \cite{ustun2019actionable}     & \xmark             & \cmark          & Binary      & \cmark               & \xmark                & \cmark          \\
\cite{wachter2017counterfactual} & \xmark             & \xmark           & Binary      & \xmark                & \xmark                & \cmark          \\

GS  \cite{laugel2017gs}    & \cmark            & \cmark          & Binary      & \xmark                & can be extended & \cmark          \\
REVISE \cite{joshi2019revise} & \xmark             & Binary       & Binary      & \xmark                &     \xmark              & \cmark          \\
CLUE & \xmark & \xmark & \cmark & \xmark & \xmark & \cmark \\
FACE  \cite{poyiadzi2020face}  & \cmark            & Binary       & Binary      &      can be extended             &        \xmark           & \xmark           \\

DiCE  \cite{mothilal2020dice} & \xmark             & \cmark          & Binary      &     \cmark (post-hoc)              & \cmark               & \cmark          \\

CRUDS \cite{downs2020cruds} & \cmark & \cmark & \xmark & \cmark & \cmark &  \cmark \\

\rowcolor{Gray} \method{}    & \cmark            & \cmark          & \cmark         & \cmark               & \cmark               & \cmark  \\\hline\hline       
\end{tabular}
\caption{Summary of existing contrastive explanation methods based on their underlying assumptions (gradient-based approaches are not model-agnostic) and on whether they handle immutable, semi-immutable  and categorical features. We also highlight methods that can generate diverse counterfactual examples (CEs) and those whose CEs are generated or selected from the training set.}\label{tbl:lit}
\end{table*}

Much of the recent work surrounding ExAI revolves around interpreting decisions in a post-hoc manner \cite{ribeiro2016why,selvaraju2017grad,patro2019u,casalicchio2018visualizing}. One can alternatively achieve explainability by matching human argumentation by contrasting models' decisions. \cite{miller2018contrastive} introduced contrastive explanations by relying on foundations in philosophy and cognitive science as two types of contrastive why-questions: alternative why–questions, addressing the ``rather than'' part, and congruent why–questions, addressing the ``but'' part.  \cite{dhurandhar2018explanations} argued that such forms can be found in many human-critical domains such as medicine and criminology and proposed a novel method that finds the contrastive perturbations with minimal change in \textit{any} black-box deep model. Their contrastive explanations method (CEM) solves an optimization problem on pertinent positive and pertinent negative inputs to alter the decision. A similar optimization is used in \cite{ustun2019actionable} to provide contrasts in linear models and in \cite{wachter2017counterfactual} through gradient descent. 
% Later, \cite{van2018contrastive} generalize this contrast by training a decision tree centered around a particular point of interest to learn the contrast between the fact and the foil as a set of human-understandable contrastive rules that represent the contrastive explanation of the model's decision. 

Prior to that, the growing spheres (GS) generative approach was proposed in \cite{laugel2017gs} to locally understand the decision boundary of a classifier with no reliance on existing training data. This decision boundary is interpreted to find the minimal change needed to alter an associated outcome. Later, \cite{joshi2019revise} develop REVISE, a counterfactual framework where the underlying data distribution is estimated and used to generate the smallest set of changes to improve an outcome on any differentiable decision-making system. Within the uncertainty estimation framework, \cite{antoran2020clue} highlight input features that are responsible for uncertainty in probabilistic models in their method CLUE.

\cite{poyiadzi2020face} argue that existing methods might yield counterfactual paths that are not practical making actionable recourse infeasible. Accordingly, the authors relied on the shortest paths defined via density-weighted metrics to provide their Feasible and Actionable Counterfactual Explanation, FACE. Other approaches include \cite{rathi2019generating} where SHAP values \cite{lundberg2017unified} are leveraged, \cite{dhurandhar2019model,pawelczyk2020learning} where explanations are model-agnostic explanations for structured datasets and \cite{van2018contrastive_rl,lucic2022focus} where the contrast of interest is designed to in policy-based reinforcement learning settings and non-differentiable tree models respectively. \cite{downs2020cruds} model the counterfactual search through a conditional subspace variational auto-encoder to extract latent features that are relevant and generate valid and actionable contrastive examples. \cite{mothilal2020dice}, on the other hand, solve an optimization problem that balances diversity and proximity to generate counterfactuals in their DiCE framework.

Those methods are surveyed by \cite{pawelczyk2021carla} in an extensive benchmarking Counterfactual And Recourse
LibrAry (CARLA). 
% While the distance of the factual to the nearest counterfactual is generally computed as the normalized $l0$ of $l1$ norm or any convex combination thereof, auto-encoders (AE) have driven rapid progress on the use of manifold-like distances in counterfactual methods. CARLA highlights the methods that use variational autoencoders (VAE) to map the intervention geometry to a lower dimensional latent space. The latter approximates the data distribution while capturing input dependencies yielding attainable and feasible counterfactuals. In this work, we leverage VAEs in the neighboring sampling process to locally contrast the decision of \textit{any} classifier. 
% We implement our method and benchmark against existing work by relying on CARLA infrastructure. 
Table~\ref{tbl:lit} summarizes the surveyed counterfactual methods by highlighting their specifications. 

\section{Methodology}\label{sec:method}
We consider a  multi-class black-box classifier $f: \mathcal{X} \mapsto [0,1]^{|C|}$ with $f(x)$ being a $c-$dimensional vector specifying the probability of $x$ belonging to each class in $C$. 
We consider an input $x, f(x) = y_{\texttt{fact}}$ for which we would like to derive a close CE $x', f(x') = y_{\texttt{contrast}}$. We define  $g: \mathcal{Z} \mapsto C$ to be an entropy-based approximation of $f$. Given a proximity measure $\pi$ and an edit distance $\delta_g$, the contrastive $x' \in \mathcal{X}$ is obtained by minimizing $\delta_g(x,x')$, and the approximation loss computed on local neighbors of $x$, $\mathcal{L}_{\pi}(f,g)$ while imposing a regularization component. An additional constraint is for the model prediction on $x'$ to be the desired contrast class. The problem can thus be formulated as:
\begin{align}\label{eq:opt}
	 \underset{x'}{\arg\min} & \hspace{1em} \mathcal{L}_{\pi_x}(f,g) + \lambda_1 R(g) + \lambda_2 \delta_g(x, x') \\
	\text{subject to } & \hspace{1em}  f(x') = y_\texttt{contrast}	
\end{align}, with $\lambda_1$ and $\lambda_2$ are regularization parameters on the complexity of $g$ and the proximity measure respectively. 

\begin{figure*}[t]
    \centering
    \includegraphics[width=0.95\textwidth]{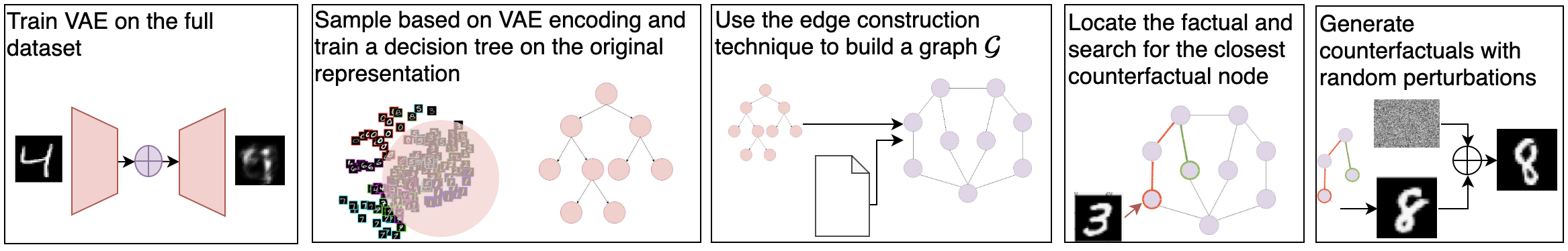}
    \caption{Overview of \method{}}
    \label{fig:method}
\end{figure*}

We refer to the approximation loss $\mathcal{L}_{\pi_x}(f,g)$ as locality-aware fidelity loss. We imply that a model $g$ minimizing the fidelity loss should yield similar outcomes as $f$ in the local neighborhood $\tilde \pi_x$. Assuming a model $g$ to be a \textit{faithful} approximation of $f$, the constraint can be replaced by $g(x') = y_\texttt{contrast}$. The model-agnosticity requirement of our approach thwarts any assumptions on $f$, thus, any gradient-based solution. Alternatively, we force the constraint by reducing our search space to nodes in the Decision Tree (DT) corresponding to $g(x') = y_\texttt{contrast}$ that minimizes the locality-aware fidelity loss $\mathcal{L}_{\pi_x}(f,g)$. We then minimize the edit distance through our one-to-many shortest path problem based on contrast boundaries learned by $g$. Consequently, we encourage feature changes with low entropy (i.e. high info gain) that can alter decisions. Our methodology is visualized in Figure~\ref{fig:method}.

\subsection{Minimizing Locality-aware Fidelity Loss}
We \textit{locally} approximate the behavior of $f$ with a DT. DTs are favored given their ability to provide a range of CEs rather than single points in the counterfactual world. Additionally, simple $g$ models are desired as they are highly interpretable but might not yield good approximations. We model this trade-off by minimizing $\mathcal{L}_{\pi}(f,g)$ while maintaining low complexity of $g$ through the regularizer $R(g)$. 
% In the context of contrastive explanations, compact models that still satisfy accuracy requirements are harder to manipulate \cite{slack2021counterfactual}.
Lastly, to satisfy the immutability requirement of some features such as gender, we remove such features from the input. % by considering a subset $\mathcal{X}_{\text{mut}} \subset \mathcal{X}$ instead. 

\subsubsection{Sampling in the local neighborhood}
Existing work computes distances between samples based on $L-p$ norms, most akin in $p=0$ (edit distance) or $p=2$ (edit L2 cost) or on domain experts to elicit the appropriate distance function \cite{mc2018interpretable,lash2017generalized}. We employ a manifold-like distance to approximate actual distances while reflecting attainability. Such non-Euclidean distance is also suitable for categorical and non-tabular data. We mainly utilize VAEs which demonstrated significant performance gains in approximating manifold-like distances. Mainly, VAEs learn a new geometry, potentially denser, of the intervention space while encoding correlations, feasibility and the plausibility of a CE occurring.

Hence, given the input $x$, we learn its latent representation $z$ in a self-supervised manner  by defining a latent model $p(x) = \int p(x | z ) p(z) dz$, an encoder $m(.)$ with its parameters $\Phi$ and distribution $q_{\Phi}(z|x)$, and a decoder $g(.)$ parameterized by $\Theta$ with a likelihood of $p_{\Theta}(x|z)$. $\Phi$ and $\Theta$ are represented by potentially non-linear functions. We then define $p_{\mathcal{D}}(x)$ as the empirical distribution of the data. Under these circumstances, the evidence lower bound (ELBO) \cite{kingma2013auto} can be used to compute the intractable integral above as:
\begin{multline}\label{eq:auto-encoders}
	\expected_{p_\mathcal{D}} (x) \bigg[\log p_{\Theta} (x)\bigg] \ge \expected_{p_\mathcal{D}} (x) \bigg[ \expected_{q\Phi(z|x)} \big[\log p_{\Theta} (x | z) \big] 	\\
	-\mathbb{K}\mathbb{L} \bigg( q_{\Phi} (z | x) || p(z)\bigg) \bigg]
\end{multline}

where $\mathbb{K}\mathbb{L}$ is the relative entropy or the Kullback–Leibler divergence \cite{kullback1951information}. Moreover, $q_{\Phi}(\bmz|\bmx)$ and $p_{\Theta}(\bmx|\bmz)$ are assumed to be Gaussian. In the case of binary attributes, the decoder can be assumed to be Bernoulli.
% The latent representation $\bm{z}$ can be found by optimizing the right-hand side of the Equation~\ref{eq:auto-encoders} above. 
Once computed, the encoding $z$ will be utilized to compute proximity as $\pi(x,x') = || z - z' ||_2$.

\subsection{Minimizing Counterfactual Cost Through Graph Search}\label{sec:dt_search}
$g$ gives a family of counterfactuals inferred from every leaf node labeled $\yfoil$. The goal is to search for the most proximate counterfactual. The search path $\yfact \leadsto \yfoil$ can be directly translated into a CE. To this end, we construct the directed weighted graph $\mathcal{G} = (\mathcal{V}, E)$ with $\mathcal{V}$ constructed from $g$ nodes where $v\in\mathcal{V}$ can be a leaf node (class label) or an internal node (decision). The goal is to reach $\yfoil$ from $\yfact$ through the path reflecting the minimal edit. 

\subsubsection{Edge construction}
We consider the edge $e_{ij} \in E$ connecting decision node $i$ to decision/label node $j$. $e_{i,j}$ represents a decision $f_i \bigoplus v$, where $f_i$ is the feature in the node $i$, $\bigoplus$ is an operator and $v$ is a threshold value or a category. $e_{i,j}$ is proportional to the edit cost of $f_i$. We assume similar costs of edges $e_{ij}$ except for the following cases: 
\begin{itemize}
    \item Custom cost function where the user specifies an edit cost $c_{\text{edit}}$ of $f_i$ (e.g. the cost of changing a job is twice that of relocating). In this case, all edges $e_{ik}$ inherit $c_{\text{edit}}$.
    \item Semi-immutability where editing feature $i$ is only possible in a particular direction (e.g \textit{has\_degree} cannot be set to \textit{false} when it is \textit{true}). In view of this, we assign an infinite weight to the corresponding edge. 
\end{itemize}
% \textcolor{red}{normalize edge weights by leaf confidence or average leaf confidence (in the case of an intermediate node)}

\subsubsection{Graph search}
The vertices $v\in\mathcal{V}$ correspond to one of the following labels: (1) \texttt{fact}, (2) \texttt{contrast}, and (3) internal decision node. We identify, $u_\text{start}$ the \textit{unique} start node from category (1) that corresponds to $x$. Then, we execute a one-to-many shortest path problem on $\mathcal{G}$ from $u_\text{start}$ to nodes in category (3). Once the search resumes, the result will be in the form of feasible rules $f_i \bigoplus v$. For practitioners who are interested in the CE instead of the contrastive path, we derive $x'$ as follows. We consider $f$ to be a feature in $x$. If $f$ is not part of the contrastive path, its value is kept intact in $x'$. Otherwise, it is altered according to the $f_i \bigoplus v$ with a margin $\sim \mathcal{N}(0, \frac{\sigma_i}{m})$ with $m$ is a tunable parameter and $\sigma_i$ is the standard deviation of $f_i$. For categorical values no random perturbations are applied. 

% Similar to existing counterfactual methods, \method{} might overlook dependencies between the covariates in the data; a CE requiring \textit{job promotion} can miss its correlation with \textit{age} and higher \textit{salary}). We wish to explore correlation integration as a future work. 

% Consequently, \method{} generates rule-based, feasible and  diverse counterfactuals. Additionally, the contrast boundaries learned by the tree are generated by sampling from a manifold to encourage plausible and attainable counterfactuals.  
\subsection{Complexity}
Lastly, we study the complexity of \method{} that is comprised of three main components: local sampling, DT training, and graph search. The former two components are extensively studied for optimizations in the literature through vector quantization \cite{avq_2020}, pre-pruning and ensembling. We focus our study on the third component which constitutes the main building block of \method{}. The problem can be cast into a single source with non-negative edge weights and no cycles; thus Dijkstra's algorithm is a suitable infrastructure. With a Fibonacci, instead of a binary, heap, the complexity can be optimized to $\mathcal{O}(E + V \log V)$ \cite{dijkstra2001oral}. Furthermore, the constrained construction of  $\mathcal{G}$ gives rise to the following guarantees: 

\begin{itemize}
    \item $|V| <<< 2^{\texttt{max\_depth}}$ is controlled by (1) the size of input that affects the \textit{max\_depth} parameter of the DT and (2) the pruning techniques that avoid over-fitting. 
    \item $|E| \le |V|-1$. Equality holds only when no semi-immutability constraints are imposed. 
\end{itemize}
Accordingly, the complexity becomes $\mathcal{O}(V\log V)$ with a bounded $|V|$ that follows from small max\_depth parameters. % imposed by the regularizer $R(g)$.

\section{Experiments}\label{sec:exp}
\begin{figure*}[t]
    \centering
    \includegraphics[width=\textwidth]{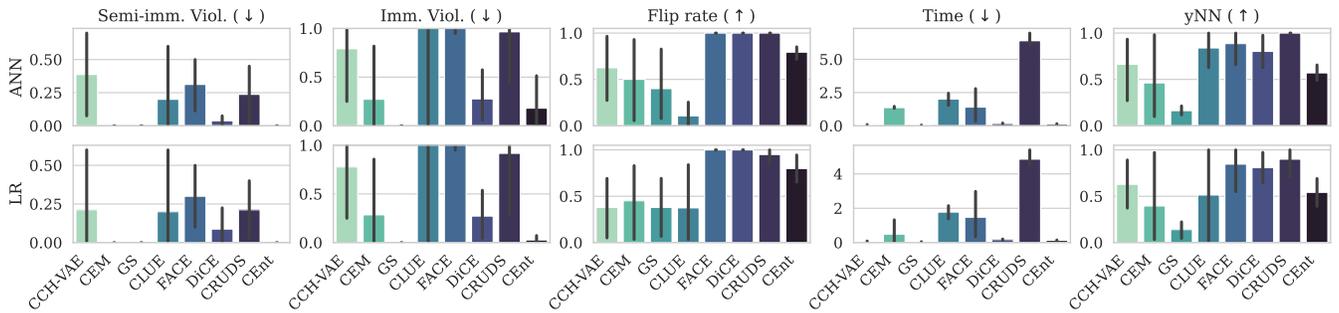}
    \caption{Results averaged on four numerical datasets}
    \label{fig:metrics_adult}
\end{figure*}

\begin{figure*}[t]
    \centering
    \begin{subfigure}{0.48\textwidth}
        \includegraphics[width=\linewidth]{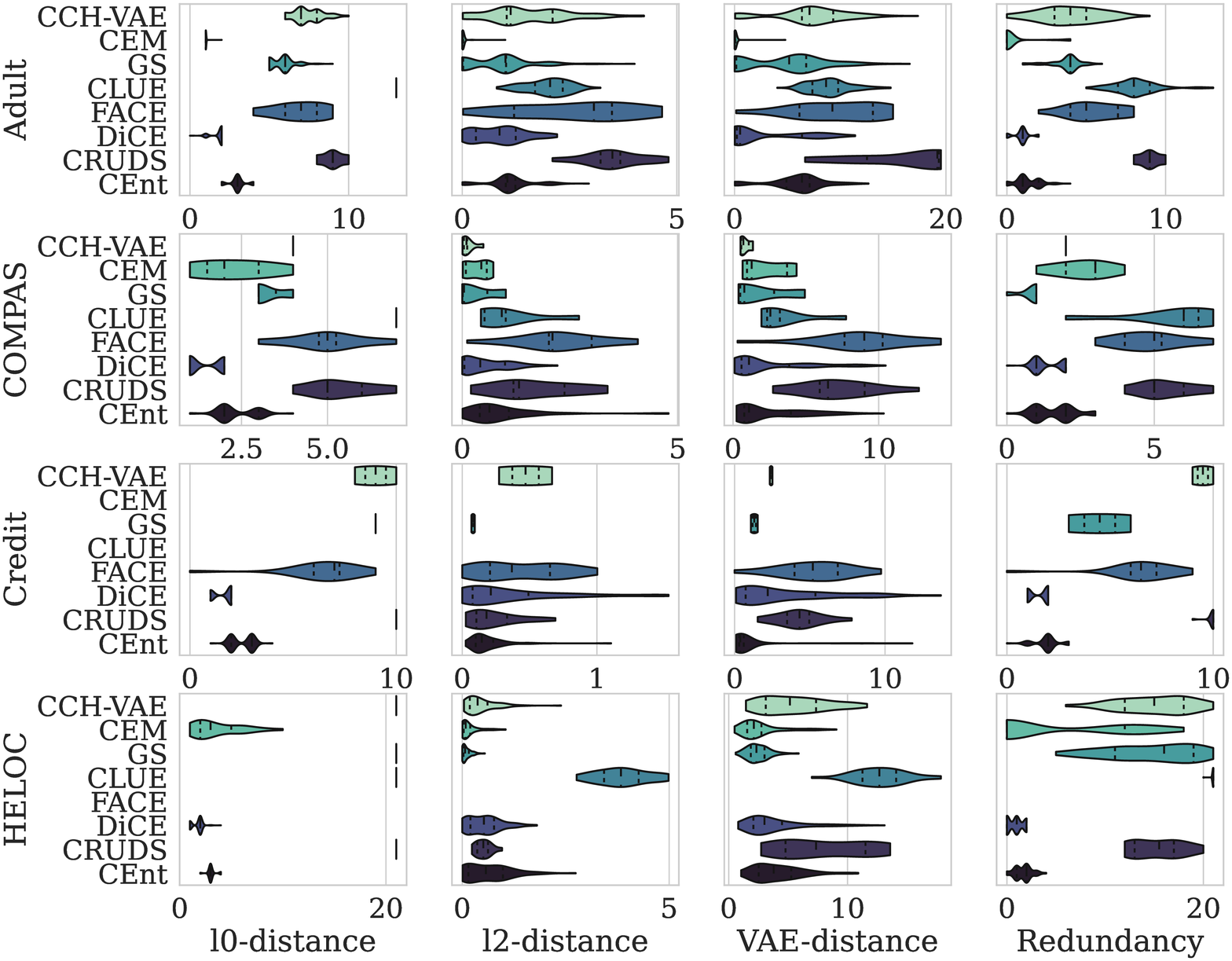}
        \caption{LR}
    \end{subfigure}
    \begin{subfigure}{0.48\textwidth}
        \includegraphics[width=\linewidth]{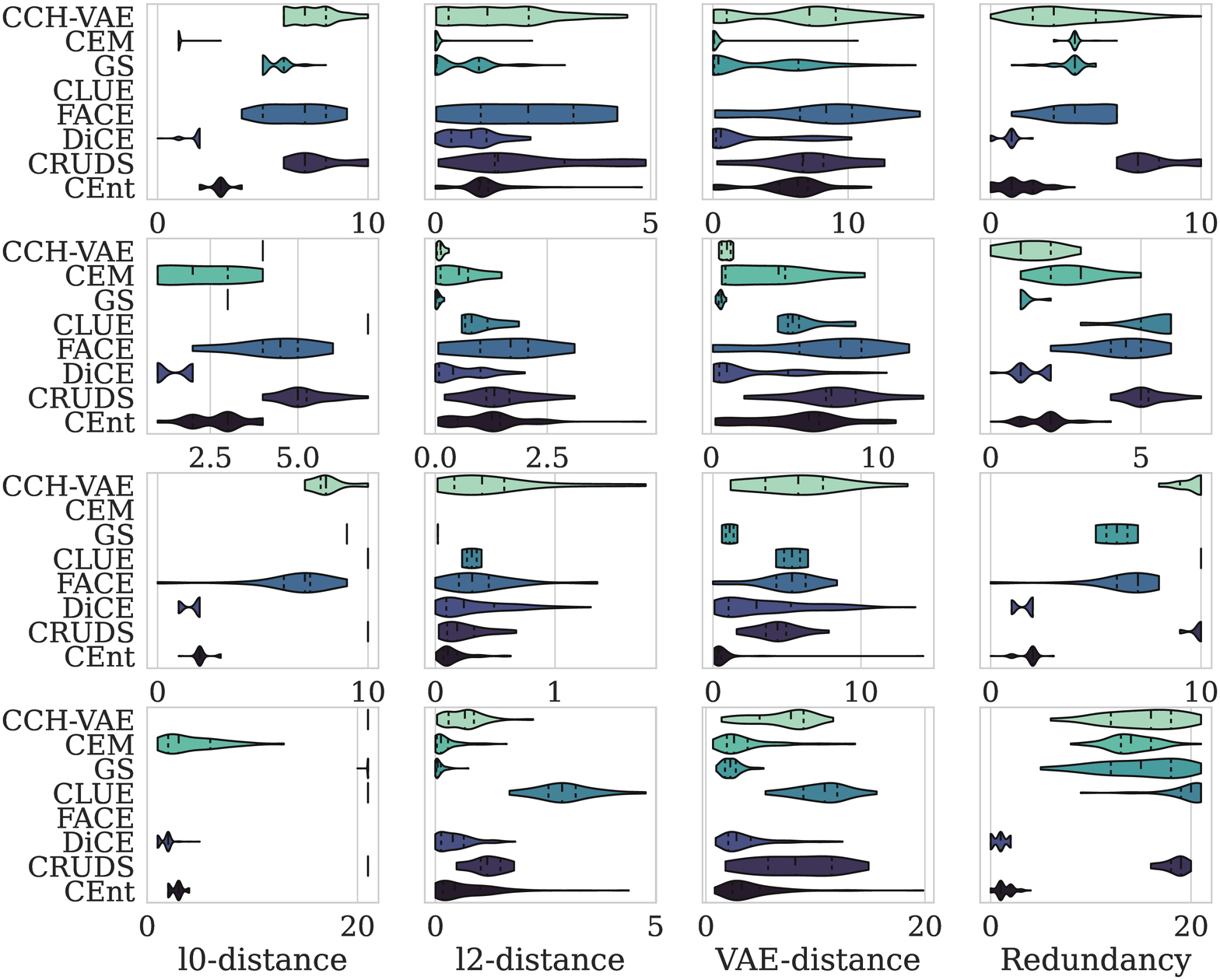}
        \caption{ANN}
    \end{subfigure}
    \caption{Distribution of proximity scores across different contrastive methods on the four numerical datasets}
    \label{fig:proximity}
\end{figure*}

In this section, we validate \method{} on a variety of datasets. We demonstrate its extension to imagery data and a special use-case for detecting vulnerabilities of textual classifiers. 
\subsection{Setup}
We implement \method{} within CARLA framework \cite{pawelczyk2021carla} which also provides an implementation of existing recourse techniques \footnote{https://github.com/carla-recourse/CARLA}. Experiments are run on 2 cores of Intel(R) Xeon(R) with 12GB RAM. We train 2 models on the numerical datasets: a logistic regression (LR) model and a neural network (NN). The NN consists of 2 layers with 13 and 4 neurons activated via \textit{relu} and trained using a weighted binary cross-entropy loss function through gradient-descent with root mean squared propagation. 

\subsubsection{Numerical Datasets}
Four numerical datasets \cite{pawelczyk2021carla} are used in this work. The Adult dataset is used to predict whether an individual has an income $>=50$K USD/year and consists of 48,842 instances and 14 attributes with \textit{age}, \textit{sex} and \textit{race} are set as immutable. COMPAS consists of information about more than 10,000 criminal defendants and is used by the jurisdiction to score the re-offending likelihood. The immutable features for COMPAS are \textit{sex} and \textit{race}. The Credit dataset consists of 150,000 attributes and 11 features to predict the possibility of financial distress within the next two years with \textit{age} being the only immutable feature. Finally, the HELOC dataset consists of 21 attributes describing anonymized information about a home equity line of credit applications made by real homeowners. HELOC has 9871 instances used to predict whether the homeowner qualifies for a line of credit or not based on 21 features with no immutability constraints. 

\subsubsection{\method{} Settings}
We train a VAE with batch normalization for $10$ epochs with a learning rate of $0.001$ and a dropout rate of $0.2$. The weight used in the KL divergence is $2.5\times 10^{_4}$. The number of hidden layers and neurons in our VAE is adaptive to the input size and is selected according to the best validation loss. For Adult data, we use one layer of 25 neurons and a bottleneck of size 8. For Credit and COMPAS, we use a layer of 16 neurons and a bottleneck of size 7 whereas for HELOC, we utilize 2 hidden layers with 25 and 16 neurons and a bottleneck of size 12. For image datasets, we employ 3 layers of 500, 250 and 32 neurons. Sampling $k$ neighbors is efficiently achieved by utilizing \cite{dong2011efficient}\footnote{https://github.com/lmcinnes/pynndescent} with k=1000 equally distributed among the \texttt{fact} and \texttt{contrast} classes.
We split our data into 80\% training used to train $f$ and 20\% testing used to test \method{} and other contrastive methods. We set the \textit{max\_search} parameter to 50, i.e. we try at most 50 diverse CEs, if none flips the prediction we claim failure to produce a counterfactual and we return the one tried at last. 

\subsubsection{Metrics}

We employ 9 metrics to assess the following aspects in the contrastive search. 

\begin{itemize}
    \item \textit{Fidelity}. We evaluate the local performance of the DT, $g$, in approximating $f$ through model accuracy with respect to $f$'s predictions. We also compute the rate of semi-immutability constraint violations (\textit{age} cannot decrease in the CE) and immutability violations. Fidelity is reflected by higher accuracies and lower violation rates. 
    \item \textit{Proximity}. We evaluate how close the derived CE $x'$ is to its original counterpart $x$. $l0$-cost computes the number of feature changes between $x$ and $x'$. $l2$-norm reflects the Euclidean distance between $x$ and $x'$ as $\sqrt{\sum_i (x_i - x'_i)^2}$. We also compute $\pi$, the $l2$-distance on the VAE encodings to reflect the manifold-like distance as $\sqrt{\sum_i (z_i - z'_i)^2}$. Finally, we compute redundancy, as in \cite{pawelczyk2021carla}, to evaluate how many of the proposed feature contrasts were not necessary. This is achieved by successive flipping operations of values in $x'$ into $x$ and inspecting whether the label would flip back. Lower distances and redundancy scores are favored.
    \item \textit{Flip rate}. We test the ability of  $x'$ in changing the prediction (a.k.a success rate). Higher scores are favored. 
    \item \textit{Latency}. We measure the time needed to derive a counterfactual in seconds. In the methods that require VAE encodings, training VAEs is excluded from latency calculation but obtaining the encodings is included. 
    \item \textit{Agreement}. We finally measure the agreement between $x'$ and its neighbors by computing the \textbf{yNN} score as in \cite{pawelczyk2021carla} with $k=5$. A score $\approx$ 1 implies that the neighborhood consists of points with the same predicted label as the CE $x'$; thus an attainable CE.
\end{itemize}

\subsection{\method{} on Numerical Datasets}

We randomly sample 100 instances from the testing data equally distributed among positive and negative labels, and we test \method{} on both models and all 4 datasets. We compare \method{} against CCH-VAE \cite{pawelczyk2020learning}, CEM \cite{dhurandhar2018cem}, GS \cite{laugel2017gs}, CLUE \cite{antoran2020clue}, FACE \cite{poyiadzi2020face}, DiCE \cite{mothilal2020dice} and CRUDS \cite{downs2020cruds}.

\begin{table}[h]
\centering
\footnotesize
\begin{tabular}{lllll}
       & \multicolumn{2}{c}{LR} & \multicolumn{2}{c}{NN} \\
       \cmidrule(lr){2-3} \cmidrule(lr){4-5}
       & $f$    & $g$ & $f$    & $g$ \\\hline
Adult  & 84 & 95  & 84 &  94 \\
COMPAS & 84 & 97  & 81 & 96  \\
Credit & 93 & 98  & 93 & 99  \\
HELOC  & 73 & 87  & 72 & 89 \\\hline
\end{tabular}
\caption{Accuracies on the original model $f$ and the DT $g$}\label{tbl:accuracies}
\end{table}

Fidelity with respect to $f$ is reflected in high $g$ scores, implying an accurate approximation, as shown in Table~\ref{tbl:accuracies}. The violation percentage, flip rate, latency, and agreement are reported in Figure~\ref{fig:metrics_adult} on the LR and ANN models averaged across datasets. \method{} consistently respects immutability and semi-immutability constraints that are significantly violated with methods such as FACE and CLUE. Results also show that \method{} can derive CEs in $<1$ sec that can successfully change the prediction with a $\sim$ 90\% probability. Finally, the yNN scores surpass CCH-VAE, CEM and GS but are lower than FACE, DiCE and CRUDS which shows competitive attainability scores. % This can be explained by the fact that \method{} relies on VAE distance while sampling neighbors and yNN scores are computed based on Euclidean measures. Figure~\ref{fig:distances} shows how l2-distance can be deemed small while VAE-distance, reflecting actual proximity, is high. 

The distribution of the average proximity metrics per model is shown in Figure~\ref{fig:proximity} across all datasets.
\method{} achieves significantly low edit distances ($l0$) in most of the cases. Similarly, $l2-$ and VAE-distances are small except in some cases where methods such as CEM and DiCE can derive closer CEs. It is worth mentioning that significantly low distance scores, such as in CEM, are an indication of an underlying failure in contrasting the prediction. In our benchmarks, such results were coupled with a success rates that can go below 40\% for CEM, CCH-VAE and GS. In such cases, derived CEs are very close to their original counterpart so that they do not flip the decision. \method{}, on the other hand, keeps searching for a counterfactual with no threshold on the distance. This led to consistently low distance measures while maintaining high flip rates. In this respect, it is not surprising that the significantly low redundancy scores attained by \method{} demonstrate the sufficient aspect of the counterfactuals in altering the model's prediction whereas methods such as CRUDS, GS, and CEM can a redundancy score of up to 20 on the HELOC dataset. This implies that most of their derived feature where unnecessary in the CE aspect. 
% \begin{figure}[h]
%     \centering
%     \includegraphics[width=0.4\textwidth]{figs/distances.eps}
%     \caption{VAE vs. Euclidean distance measures}
%     \label{fig:distances}
% \end{figure}

\subsection{Derivation of Visual Contrasts}
\begin{figure}[h]
    \centering
    \begin{subfigure}{0.48\textwidth}
        \centering
        \includegraphics[width=0.95\textwidth]{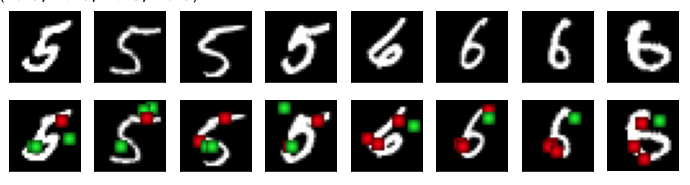}
        \caption{5 vs. 6}
    \end{subfigure}
    \begin{subfigure}{0.48\textwidth}
        \centering
        \includegraphics[width=0.95\textwidth]{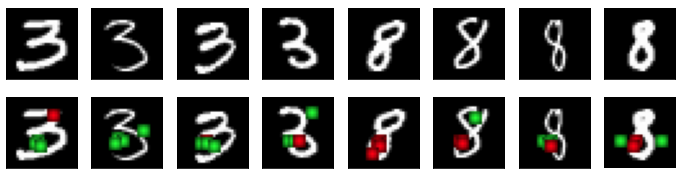}
        \caption{3 vs. 8}\label{fig:mnist38}
    \end{subfigure}
    \begin{subfigure}{0.48\textwidth}
        \centering
        \includegraphics[width=0.95\textwidth]{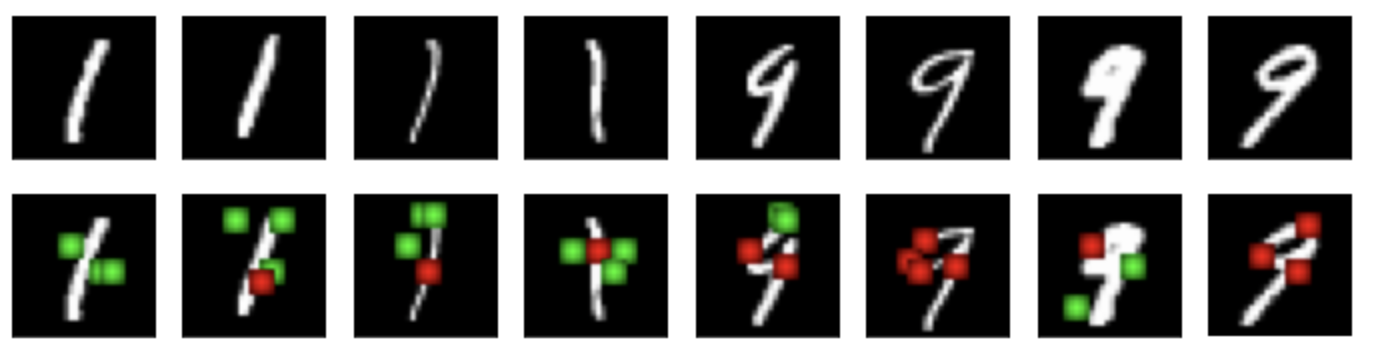}
        \caption{1 vs. 9}
    \end{subfigure}
    \caption{Visual contrast with \method{} on MNIST (red represents PPs and green represents PNs) }
    \label{fig:mnist}
\end{figure}

We consider the handwritten digit recognition, MNIST, dataset \cite{deng2012mnist} consisting of 60,000 28x28 images. We consider a pixel intensity to be a feature and we derive CEs for the binary classification of confusing digit pairs, i.e. 5 vs. 6, 3 vs. 8, and 1 vs. 9. To this end, we train a 
CNN with a convolutional layer with 28 units and a 3x3 kernel followed by max pooling and a dense layer of 128 neurons and \textit{relu} activation. A dropout of 0.2 is applied and the activation at the output is \textit{softmax}. We define the visual contrast to be a Gaussian kernel around a pixel whose intensity changed in 
$x'$. If the intensity is amplified in $x'$, the contrast is a pertinent negative (PN) whereas it is a pertinent positive (PP) if the intensity is reduced in $x'$. The visual contrasts on 8 random images for each pair are visualized in Figure~\ref{fig:mnist}. Generally, for the $5\rightarrow 6$ contrast, PNs are the pixels that close the left corner in the lower curve of number 5 and those that make its upper part more curved. PPs are mostly concerned with the upper part of 5. PPs and PNs are reciprocated with the  $6\rightarrow 5$ contrast. More importantly, the derived contrasts are mostly sufficient; i.e. no redundant pixels have been derived an exception in the third image where an outlier region is highlighted in the upper left corner. Interestingly, the $3\rightarrow 8$ is mostly related to the lower curves and not the upper one. This can be explained by the inconsistency in closing the upper loop in 8 (fifth example in Figure~\ref{fig:mnist38}) whereas the lower one is almost always closed. Similarly, CNN mostly attends to the curve of $9$ in the contrast $1\rightarrow 9$. It is worth mentioning that, in some rare cases, \method{}  fails to find a CE; i.e. it derives a CE that does not flip the model's decision. However, in such cases, the contrastive path was, intriguingly, a reasonable and visually appealing contrast even if it did not suffice to change the model's decision. 

\begin{figure}[h]
    \centering
    \includegraphics[width=0.4\textwidth]{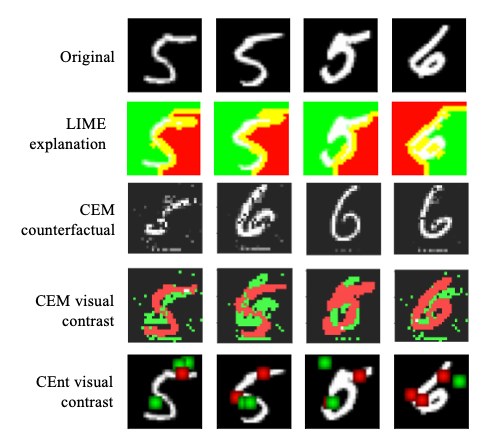}
    \caption{mnist explanations (LIME highlights pixel relevance where red is positive and green is negative, CEM and \method{} highlight contrasts where red is PP and green is PN)}
    \label{fig:cem}
\end{figure}

We select CEM and LIME for a qualitative comparison. Comparison against other methods was not possible given their design tailored for tabular datasets. General methods such as FACE and GS are also not compatible with \method{} as the former reports a series of successive examples to reach a CE and the latter's results were not reproducible. Predominantly, explanations derived by LIME, in Figure~\ref{fig:cem}, were not useful on MNIST. This can be attributed to the non-contrastive aspect of LIME and its dynamics of operating on super-pixels. The latter reason is crucial; LIME relies on segmentation techniques and is not designed to derive visual contrasts. Additionally, the first three CEs derived by CEM are visually appealing and the last one does not succeed in changing the prediction. However, the visual contrast is not obvious where CEM has the tendency to reshape the digit while flipping a great deal of pixels. \method{} is a minimally invasive process that highlights sufficient visual contrasts without major mutation on the shape.

% \begin{figure}
%     \centering
%     \includegraphics[width=0.49\textwidth]{figs/mnist_cent.png}
%     \caption{\method{} on  MNIST  (red is PP and green is PN) }
%     \label{fig:mnist}
% \end{figure}

\begin{figure}[h]
    \centering
    \begin{subfigure}{0.48\textwidth}
        \centering
        \includegraphics[width=0.95\textwidth]{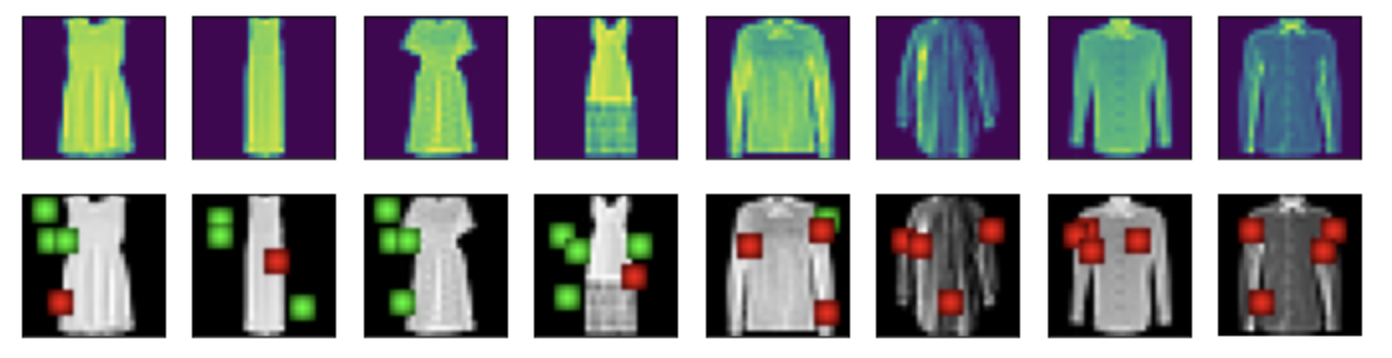}
        \caption{Dress vs. shirt}
    \end{subfigure}
    \begin{subfigure}{0.48\textwidth}
        \centering
        \includegraphics[width=0.95\textwidth]{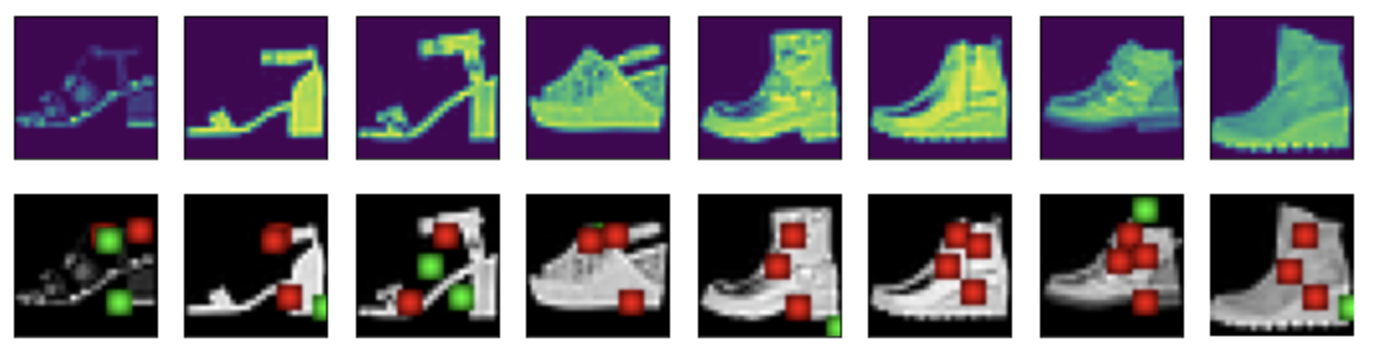}
        \caption{Sandal vs. ankle boot}
    \end{subfigure}
    \begin{subfigure}{0.48\textwidth}
        \centering
        \includegraphics[width=0.95\textwidth]{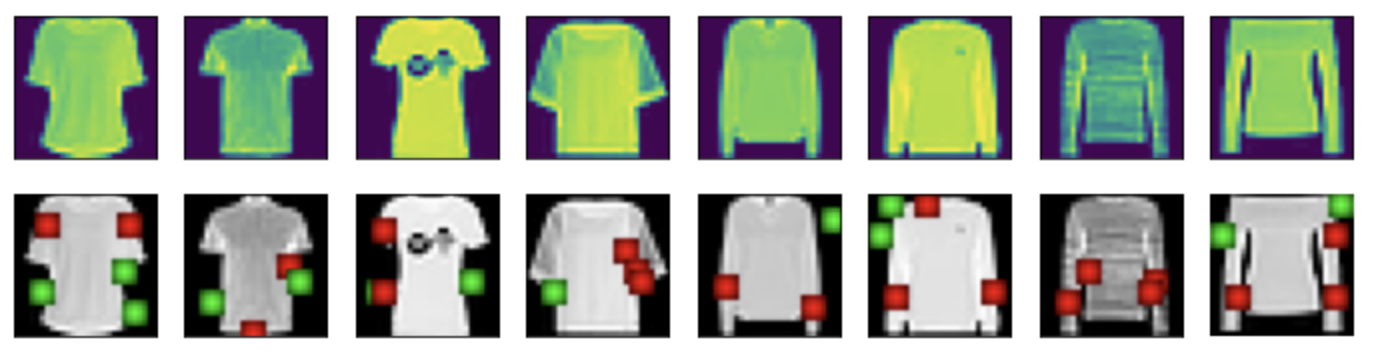}
        \caption{T-shirt vs. pullover}
    \end{subfigure}
    \caption{Visual contrast with \method{} on Fashion MNIST (red represents PPs and green represents PNs) }
    \label{fig:fashion_mnist}
\end{figure}

We also showcase \method{} on the fashion MNIST dataset consisting of 10 different cloth labels for 70,000 28x28 images. We choose 3 pairs of classes that are likely to be mistaken: dress vs. shirt, sandal vs. ankle boot, and T-shirt vs. pullover. We train the same CNN as in the MNIST case and we derive visual contrasts on 8 randomly chosen images in each category in Figure\ref{fig:fashion_mnist}. 
The dress vs. shirt contrast is mostly concerned with the sleeves as well as the width of the object. Since dresses are taller than shirts, fitting both in a 28x28 image makes the shirts wider. Intriguingly, \method{} detects this contrast. For sandal vs. ankle boot, \method{} highlights the open holes as PNs in sandals and the compact areas as PPs in the boots. Finally, \method{} detects the absence/presence of sleeves when contrasting the T-shirt to the pullover. 
\subsection{Textual Vulnerabilities Detection}
\begin{figure}[h]
    \centering
    \includegraphics[width=0.45\textwidth]{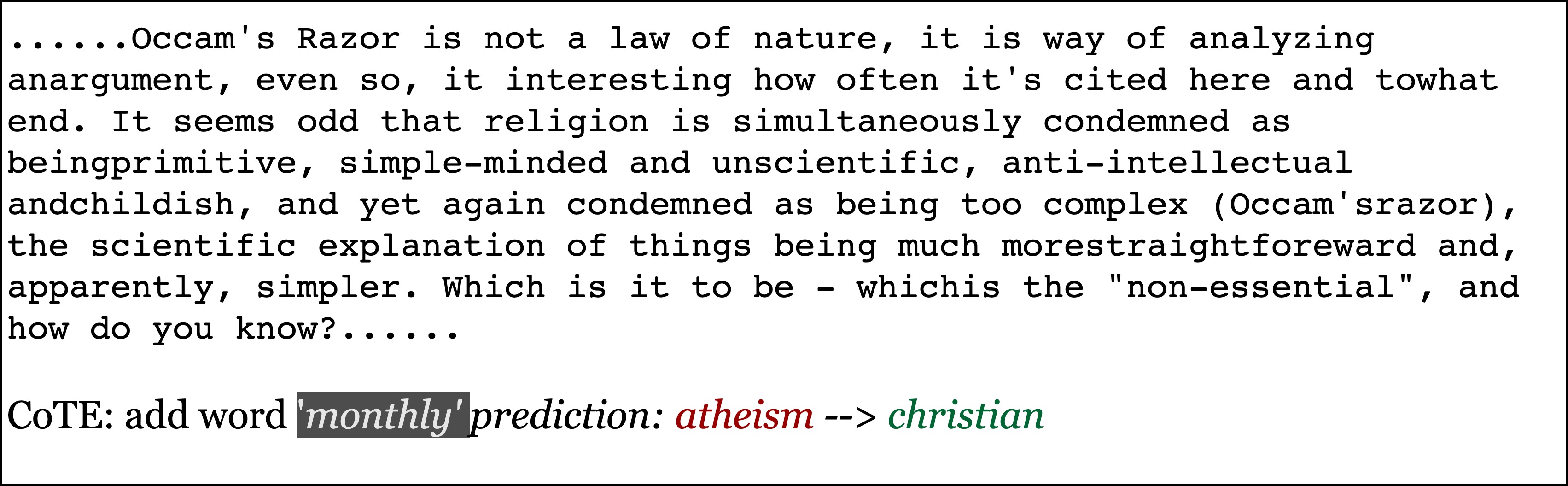}
    \caption{\method{} on an instance of the 20 newsgroup data}
    \label{fig:ex_nlp}
\end{figure}

We study an interesting use-case of \method{} in detecting non useful CEs that serve as adversarial attacks. Instead of employing a VAE distance, we consider a bag-of-words approach (BoW) where sentences are deemed close based on their words with no context, syntax, or semantic integration. Four classifiers were trained on the 20 newsgroup dataset: a random forest, logistic regression, SVM, and a neural network with two fully-connected layers consisting of 100 and 50 neurons. The approximation $g$ achieved an accuracy of 98, 93, 96, and 99\% respectively. Remarkably, the length of the contrast is on average 1 in all models implying that an insertion or deletion of exactly one word would change the model's prediction. While shorter lengths are an indication of more concise explanations, they do not necessarily reflect the same for textual data when BoW representations. For instance, as displayed in Figure~\ref{fig:ex_nlp}, including the word ``monthly'' would change the prediction from \textit{atheism} to \textit{christian} showing the sensitivity to particular words. In this case, \method{} serves as a debugging tool that highlights vulnerabilities in the context of adversarial attacks.  

\section{Conclusion}\label{sec:conc}
In this work, we develop a wide plan of attack for algorithmic recourse by accounting for custom costs and elegantly addressing semi-immutability and plausibility. We propose \method{}, a novel entropy-based method, that supports an individual facing an undesirable outcome under a decision-making system with a set of actionable alternatives to improve their outcome. \method{} samples from the latent space learned by VAEs and builds a decision tree augmented with feasibility constraints. Graph search techniques are then employed to find a compact set of feasible feature tweaks that can alter the model's decision. 

Our empirical evaluation on real-life datasets shows improvement in proximity, latency, and attainability with no constraint violation. \method{} is of great potential in adapting to non-tabular data where it identifies visual contrasts and serves as a debugging tool to detect the model's vulnerabilities. These results motivate the exploration of more complicated representations, such as embeddings or super-pixels, to improve the robustness of \method{} on different data types. We also wish to study the possibility of training numerous decision trees in different input sub-spaces to alleviate the need to access training data; improving thus privacy guarantees.

\appendix 
% \section{Appendix}
% ff
\clearpage
\bibliography{refs}
\end{document}